%% file: main.tex
\documentclass[10pt,twocolumn,letterpaper]{article}

\usepackage{iccv}
\usepackage{times}
\usepackage{epsfig}
\usepackage{graphicx}
\usepackage{amsmath}
\usepackage{amssymb}

\usepackage{booktabs}
\input{definitions}

\usepackage[pagebackref=true,breaklinks=true,letterpaper=true,colorlinks,bookmarks=false]{hyperref}

\iccvfinalcopy 

\def\iccvPaperID{***} 

\begin{document}

\title{Interpret Vision Transformers as ConvNets with Dynamic Convolutions\vspace{-5pt}}

\author[1]{Chong Zhou}
\author[1]{Chen Change Loy}
\author[2]{Bo Dai}
\affil[1]{S-Lab, Nanyang Technological University}
\affil[2]{Shanghai AI Laboratory}

\maketitle

\input{sections/abstract}

\input{sections/introduction}
\input{sections/related}
\input{sections/unify}

\input{sections/rethink}
\input{sections/discussion}

\appendix
\input{sections/supp}

{\small
\bibliographystyle{ieee_fullname}
\bibliography{egbib}
}

\end{document}

%% file: definitions.tex
\usepackage{multirow}
\usepackage[dvipsnames]{xcolor}
\usepackage{enumitem}
\usepackage[ruled]{algorithm2e}
\usepackage[noend]{algpseudocode}
\usepackage{tabularx}
\usepackage{pifont}
\usepackage{subcaption}
\usepackage{authblk}

\newcommand{\inlinesection}[1]{\noindent \textbf{#1}$\,\,\,$}

\def\ph{$\cdot$}

\newcommand{\cmark}{\ding{51}}
\newcommand{\xmark}{\ding{55}}



%% file: sections/abstract.tex
\begin{abstract}
There has been a debate about the superiority between vision Transformers and ConvNets, serving as the backbone of computer vision models.
Although they are usually considered as two completely different architectures, in this paper, we interpret vision Transformers as ConvNets with dynamic convolutions, which enables us to characterize existing Transformers and dynamic ConvNets in a unified framework and compare their design choices side by side.
In addition, our interpretation can also guide the network design as researchers now can consider vision Transformers from the design space of ConvNets and vice versa.
We demonstrate such potential through two specific studies. First, we inspect the role of softmax in vision Transformers as the activation function and find it can be replaced by commonly used ConvNets modules, such as ReLU and Layer Normalization, which results in a faster convergence rate and better performance. Second, following the design of depth-wise convolution, we create a corresponding depth-wise vision Transformer that is more efficient with comparable performance.
The potential of the proposed unified interpretation is not limited to the given examples and we hope it can inspire the community and give rise to more advanced network architectures.
\end{abstract}

%% file: sections/introduction.tex
\section{Introduction}
\label{sec:intro}
\input{figures/fig-teaser}

Since 2012, when AlexNet~\cite{alexnet} excelled in the ImageNet challenge~\cite{imagenet}, convolutional neural networks (ConvNets) start to attract tremendous attention in the field of artificial intelligence and quickly replace traditional methods and handcrafted features in many tasks.
Representative milestones, such as VGGNet~\cite{vggnet} and ResNet~\cite{resnet}, have served as the de facto backbones for a long time.
In recent years, Transformer~\cite{transformer} has emerged as a new type of neural network beyond ConvNets, gradually becoming dominant in both natural language processing (NLP) and computer vision with representative works such as Vision Transformer (ViT)~\cite{vit},
Swin Transformer~\cite{swin}, DETR~\cite{detr}, and SETR~\cite{setr}.
Given such a trend, there are also works trying to close up the gaps between ConvNets and vision Transformers, proving these two types of neural networks are competitive with each other. 
Successful practices along this line include ConvNeXt~\cite{convnext} that adopts design choices from vision Transformers and RepLKNet~\cite{RepLKNet} that replaces standard convolutional kernels with larger depth-wise ones.

Nevertheless, to date, the community still continuously debates the superiority between the ConvNet-based models and the Transfomer-based counterparts.
In this paper, we equivalently convert vision Transformers to ConvNets and bring these two types of neural networks into a unified framework.

This equivalence is established by rewriting self-attention, the core building block of Transformers, as a ConvNet block consisting of static and dynamic convolutions.
Specifically, as shown in Figure~\ref{fig:teaser}, we regard the \emph{query}-\emph{key} and \emph{attention}-\emph{value} matrix multiplications as two $1\times1$ dynamic convolutions, respectively, and consider the scaled softmax operation as an activation function. 
In this way, the corresponding ConvNet block of the self-attention block is \textit{static conv.} $\rightarrow$ \textit{dynamic conv.} $\rightarrow$ \textit{activation} $\rightarrow$ \textit{dynamic conv.} $\rightarrow$ \textit{static conv.}.
Moreover, this equivalence holds for not only the vanilla ViT but also its subsequent variants, such as Swin Transformer~\cite{swin}. Therefore, as shown in Figure~\ref{fig:unify}, we select several popular Transformers and dynamic ConvNets and characterize their design choices in the proposed unified perspective.

Being able to compare design choices, which were previously considered tied to either Transformers or ConvNets, side by side, we now can rethink design choices of vision Transformers in the design space of ConvNets and vice versa. We believe this can help researchers switch to a different angle when designing a new network architecture. And designs that are proven to be successful on ConvNets or Transformers can be seamlessly transferred to each other and potentially obtain a similar gain.
To demonstrate such potential with practical examples, in this paper, we further conduct two empirical studies.

First, we investigate the role of softmax in vision Transformers as the activation function in the position of ConvNets.
On one hand, as an S-shaped function, modern ConvNets rarely adopt softmax as the activation function since studies~\cite{alexnet} show that non-S-shaped functions, such as ReLU~\cite{relu}, possess significant advantages. On the other hand, apart from being S-shaped and providing non-linearity, softmax also delivers normalization and channel-wise communication to its input features.
Therefore, we examine which role of softmax is more important in Transformers and whether it can be replaced by techniques that are dedicated to a single role and commonly used in ConvNets through ablation studies. Experiments show that non-linearity and normalization play more important parts but can be substituted by techniques other than softmax. For instance, by replacing softmax with LayerNorm~\cite{layer-norm} and ReLU~\cite{relu}, without introducing any computational cost, the top-1 ImageNet accuracy of Swin-Tiny is improved from 81.13 to 81.56. In addition, being S-shaped is a downside of softmax as it slows down the convergence rate compared to non-S-shaped functions, such as ReLU, which aligns with previous studies~\cite{alexnet} on ConvNets.

Second, when summarizing existing Transformers and dynamic ConvNets in the unified framework, we find that the mechanism of the dynamic convolutions in self-attention is quite unique, \emph{e.g.,} kernels generated from one spatial location will be shared with other locations through a global kernel bank, which is more efficient than generating a new set of kernels for each spatial location. Meanwhile, to encourage efficiency, depth-wise convolutions are widely adopted in ConvNets. Thus, we explore whether combining these two design choices can result in an even more efficient network. In specific, we keep the kernel bank mechanism of Transformers and replace the two dynamic convolutions with a single depth-wise one. As a result, we obtain a lightweight vision Transformer with comparable performance.

We consider the aforementioned studies as two examples to conduct exploration in the proposed unified framework of vision Transformers and ConvNets, but its potential is not limited to the given examples. In fact, we hope the community can be inspired by our interpretation of vision Transformers and bring up more interesting findings.

%% file: figures/fig-teaser.tex
\begin{figure}
\centering
\includegraphics[width=0.47\textwidth]{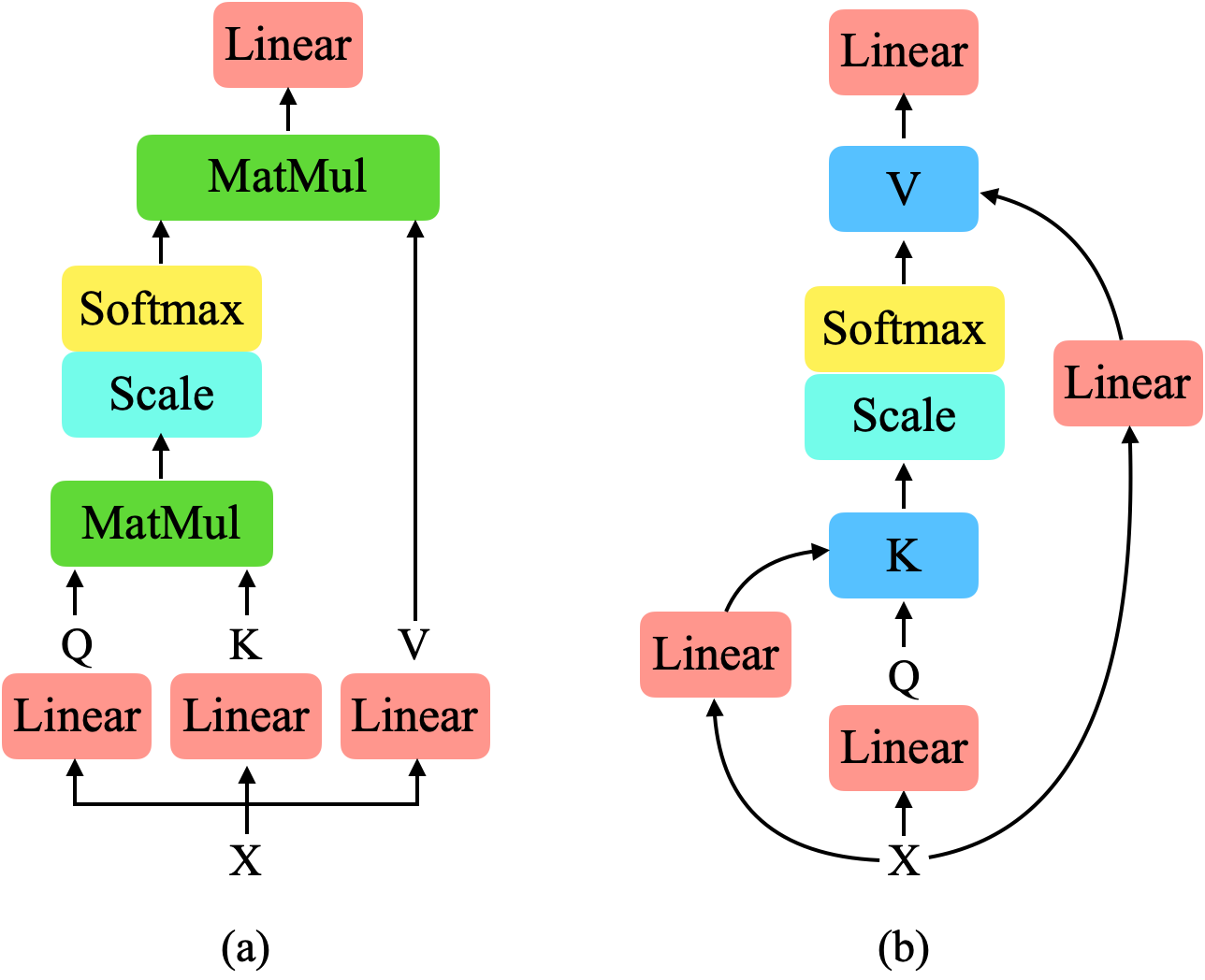}
\caption{\inlinesection{Interpret self-attention as dynamic convolutions.} The left shows a standard self-attention block. On the right, we re-organize certain modules and interpret the \emph{query}-\emph{key} and \emph{attention}-\emph{value} matrix multiplications as two $1\times1$ dynamic convolutions, respectively. Note that the two architectures are essentially identical and can be equivalently converted to each other.}
\label{fig:teaser}
\end{figure}

%% file: sections/related.tex
\section{Related Work}
\label{sec:related}
\inlinesection{Vision Transformers.} Transformers~\cite{transformer} are originally applied in NLP and are first introduced into visual recognition by Vision Transformer (ViT)~\cite{vit}. Although ViT has fewer inductive biases compared to ConvNets, which theoretically leads to better generalization ability, its quadratic computational complexity with respect to the input resolution makes it prohibitive to serve as a general backbone for computer vision tasks. Moreover, the vanilla ViT usually takes longer to converge during training. Therefore, efforts have been made to bring ConvNets modules~\cite{cvt,ceit,integration} or designs, such as feature pyramid~\cite{pyramid} and local windows~\cite{swin}, back to ViT. 
However, such integration still considers ViTs and ConvNets as two heterogeneous architectures, while we bring them into a unified framework.

\inlinesection{Dynamic convolutions.} Dynamic convolutions refer to convolutions whose kernel weights are conditioned on the inputs. Since directly generating the weights of a standard convolution is expensive, some works opt to generate weights of depth-wise convolution~\cite{d-dw-conv}. Other works maintain a learnable kernel bank and predict the coefficients followed by linear combining kernels in the bank with the predicted coefficients~\cite{d-conv}. To further cut down the computation cost, some works generate a single set of kernels for the whole feature map~\cite{senet,filter,d-conv} instead of one kernel set for each spatial location~\cite{involution,genet,d-dw-conv}. With our interpretation, we find that vision Transformers not only share commonality with existing dynamic convolutions but also contain unique designs for efficiency (\eg, implicitly approximate a dense kernel with a pair of lightweight kernels).

\inlinesection{Connections between Transformers and ConvNets.} Tay~\etal~\cite{nlp2} and Wu~\etal~\cite{nlp1} empirically compare dynamic depth-wise ConvNets with Transformers in NLP tasks. Andreoli~\cite{tensor-form} classifies convolution into index-based convolution (standard convolution) and content-based convolution whose weights are generated from input, such as the query-key matrix multiplication (attention). In this way, self-attention functions the same as content-based convolution. Cordonnier~\etal~\cite{relation} further shows that with sufficient heads and each head paying attention to a specific shift, self-attention with relative positional encoding can simulate any index-based convolutions. Han~\etal~\cite{d-dw-conv} interprets the matrix multiplication of attention and value in local vision Transformers as dynamic depth-wise convolution. Our work differs from previous works in (1) we make no assumption on the value of attention or forms of positional embedding; (2) our conversion is equivalent and bi-directional; (3) our interpretation is not limited to special variants of vision Transformers. Therefore, our work makes a further step towards practical usage and generality.

%% file: sections/unify.tex
\section{A Unified Perspective}
\label{sec:unify}
To bring vision Transformers and ConvNets into a unified framework, we focus on equivalently converting the self-attention block in Transformers into dynamic convolutions since the rest of the Transformer modules are already considered as ConvNets. In this section, we start with a brief introduction of the self-attention block (Section~\ref{sec:self-attn}), followed by a new perspective to interpret self-attention (Section~\ref{sec:convert}). Finally, we fit several existing Transformers and ConvNets architectures into the proposed framework and provide an analysis of their design choices in a unified perspective (Section~\ref{sec:existing}).

\subsection{Preliminary: Self-Attention}
\label{sec:self-attn}
Before being fed into the first self-attention block of the vision Transformer, input images are divided into patch tokens by a convolutional stem, then these patch tokens are flattened along the spatial dimension. As a result, the input to self-attention can be denoted as $x \in \mathbb{R}^{N \times C}$, where $N$ and $C$ are the numbers of tokens and embedding dimensions, respectively. Self-attention first projects $x$ to query $q$, key $k$, and value $v$ with three individual linear layers:
\begin{align}
    q = W_q x + b_q,\ k = W_k x + b_k,\ v = W_v x + b_v,
\end{align}
where $W_*$ and $b_*$ denote weight and bias. Then $q,k,v$ are divided along the embedding dimension into $H$ groups, and each group is processed separately before a final concatenation of results from each group. The mechanism is termed multi-head with each head referring to each group. Here, we demonstrate the operation for a single head:
\begin{align}\label{eq:attn}
    o_{i} = \text{softmax}(\frac{q_{i}k^\mathsf{T}}{\sqrt{C_h}})v,
\end{align}
where $*_{i}$ denotes the $i$-th token, $C_h$ is embedding dimension of each head, and $o$ is the output. 

\input{figures/fig-pair}

\subsection{View Self-Attention as Dynamic Convolutions}
\label{sec:convert}
Let us revisit Equation~\ref{eq:attn}, which represents the core of self-attention. From a general view, it can be divided into two linear matrix multiplications with a non-linear activation function (scaled-softmax) in between. When looking more closely, we observe that the output of the $i$-th token is a function of the $i$-th query and all the keys and values. In other words, keys and values are shared across queries at different spatial locations. This reminds us of a typical linear operation whose kernels are also shared across various spatial locations, convolution.

The kernel of a standard convolution can be defined as $u \in \mathbb{R}^{H \times W \times C_{\text{in}} \times C_{\text{out}}}$, where $H,W,C_\text{in},C_\text{out}$ denote the height, width, input and output channels of the kernel, respectively. The kernel slides along the spatial dimension of the input feature map and at each sliding window, the kernel performs a $(H\mathsf{x}W\mathsf{x}C_{\text{in}} \rightarrow C_{\text{out}})$ linear projection.

We find that each of the two matrix multiplications in self-attention, as shown in Figure~\ref{fig:teaser}, essentially operates in the same way as a $1\times1$ convolution. First, the \emph{query}-\emph{key} multiplication can be viewed as a convolution kernel $k \in \mathbb{R}^{1 \times 1 \times C_{\text{in}} \times N}$ sliding on the query feature map, where $N$ denotes the number of tokens. Similarly, the \emph{attention}-\emph{value} multiplication is also a $1\times1$ convolution with kernel $v \in \mathbb{R}^{1 \times 1 \times N \times C_{\text{out}}}$ and takes the attention map as the input.

\input{figures/fig-unify}
\input{figures/tables/tab-unify}

However, different from the standard convolution, the weights of kernel $k$ and $v$ are conditioned on the input therefore the converted convolutions are dynamic convolutions. In addition, we observe that kernels $k$ and $v$ operate in pairs as shown in Figure~\ref{fig:pair}. Specifically, at each sliding window, $k$ performs $N \mathsf{x}\ (1\mathsf{x}1\mathsf{x}C_{\text{in}} \rightarrow 1)$ linear projections. The resulting $N$-dimensional vector then goes through a scaled softmax activation function. Finally, kernel $v$ projects each of the $N$ scalars in the vector to a $C_{\text{out}}$-dimension vector, followed by the summation of all the $N$ resulting $C_{\text{out}}$-dimensional vectors. Overall, self-attention achieves the $(C_{\text{in}} \rightarrow C_{\text{out}})$ mapping with two convolutions instead of one, and later we will provide an analysis of how such a breakdown contributes to the efficiency of Transformers by comparing it with common practices in existing dynamic convolutions.

\subsection{Unified Framework for Existing Networks}
\label{sec:existing}
So far, our conversion is based on the vanilla ViT. Now we generalize the conversion to more Transformer architectures. In particular, apart from the ViT~\cite{vit}, we choose the Swin Transformer~\cite{swin}, which is a seminal work in the local vision Transformer family, and Linformer~\cite{linformer}, which is a representative work in Transformers with linear complexity. Besides, we also select two typical ConvNets with dynamic convolutions~\cite{d-conv,d-dw-conv} and bring all the chosen Transformers and ConvNets into a general framework so that we can analyze their design choices from a unified perspective.

As shown in Figure~\ref{fig:unify} and Table~\ref{tab:unify}, we characterize ViT, Swin Transformer, Linformer, Dynamic Convolution (D-Conv), and Dynamic Depth-wise Convolution (D-DW-Conv) from four dimensions, namely, kernel bank, kernel selection, kernel type, and kernel size.

\inlinesection{Kernel bank.} Sharing the same set of kernels across spatial dimensions contributes significantly to the parameter efficiency of convolution. However, kernel sharing is not always adopted in current dynamic convolutions, \eg, D-DW-Conv opts to generate a dedicated kernel set for each spatial location at the cost of consuming more computation.
In contrast, vision Transformers and D-Conv re-use kernels with the help of kernel banks. Specifically, D-Conv maintains a learnable kernel bank during training, and kernels in the bank serve as the basis or prototypes, which are linearly combined into a single kernel for each input image. During inference, the kernel bank stays static and is shared by all testing images. Vision Transformers, however, share kernels in a more fine-grained way, where kernels in the bank are not shared across images but across spatial locations within a single image. In particular, for each image, the kernel bank consists of kernels generated at all spatial locations. To fetch kernels from the bank, different Transformers implement different rules, which we will elaborate in the next paragraph. As the kernel bank of Transformers is conditioned on input images, we consider it as the dynamic kernel bank as opposed to the static kernel bank of D-Conv.

\inlinesection{Kernel selection.} As shown in Figure~\ref{fig:unify}, under the proposed perspective, ViT, Swin Transformer, and Linformer, as representative methods for global, local, and linear Transformers, respectively, only differ in the rule of selecting kernels from the bank. In particular, ViT selects all the kernels from the bank regardless of the input location, and Swin Transformer chooses kernels generated within the local windows. In contrast, Linformer performs soft selection, where all kernels are fed into a tiny network and result in a smaller set of kernels therefore each resulting kernel is essentially a learned linear combination of the input kernels. Apart from Transformers, D-Conv adopts kernel bank as well and it also follows a soft selection rule. But different from Linformer, D-Conv explicitly predicts the coefficients of the linear combination from the input feature. 
Moreover, we find that the softmax in the self-attention of Transformers also implicitly affects the kernel selection by biasing towards kernels that are more close to the input feature in terms of cosine similarity and we will conduct more comprehensive studies in the role of softmax in Section~\ref{sec:softmax}.

\inlinesection{Kernel type.} Directly generating the weights of a standard convolution kernel can be expensive and even infeasible. For instance, to generate kernel weights from a $C$-dimensional vector, a one-layer weights generator requires $(C\mathsf{x}H\mathsf{x}W\mathsf{x}C_{\text{in}}\mathsf{x}C_{\text{out}})$ parameters. Therefore, all the listed Transformers and ConvNets avoid direct weights generation in various ways. D-DW-Conv adopts a straightforward solution where the standard convolution is replaced with a depth-wise one (sparse channel-wise connectivity). Consequently, the required parameters are reduced to $(C\mathsf{x}H\mathsf{x}W\mathsf{x}C_{\text{in}})$. D-Conv re-parameterizes the generation into the linear combination of kernels in a static bank, where only the lightweight coefficients are dynamically predicted from the input. Different from dynamic ConvNets, self-attention in vision Transformers achieves both dense channel-wise connectivity and a dynamic kernel bank. First, self-attention adopts only $1\times1$ convolutions since it does not rely on large kernel size to gather information from distant locations, instead, it achieves spatial-wise communication by using kernels generated by each other. Second, instead of performing $(C_{\text{in}} \rightarrow C_{\text{out}})$ at once, self-attention breaks the operation into two steps, $(C_{\text{in}} \rightarrow 1)$ followed by $(1 \rightarrow C_{\text{out}})$. In this way, self-attention cuts down the required parameters of the weights generator to $(CC_{\text{in}}+CC_{\text{out}})$.

\inlinesection{Kernel size.} In the previous paragraph, we mention that self-attention adopts only $1\times1$ convolutions, which is uncommon in dynamic convolutions. This design is a double-edged sword because, on the one hand, it significantly encourages parameter and computational efficiency, and on the other hand, $1\times1$ convolutions do not embed any spatial priors, while spatial priors are beneficial to vision tasks. Therefore, vision Transformers are usually augmented with positional embedding to mitigate this shortcoming. However, it might be worth exploring whether allowing self-attention to generate larger kernels, \eg, $3\times 3$, also addresses the problem. We leave it as a future work.

%% file: figures/fig-pair.tex
\begin{figure}
\centering
\includegraphics[width=0.3\textwidth]{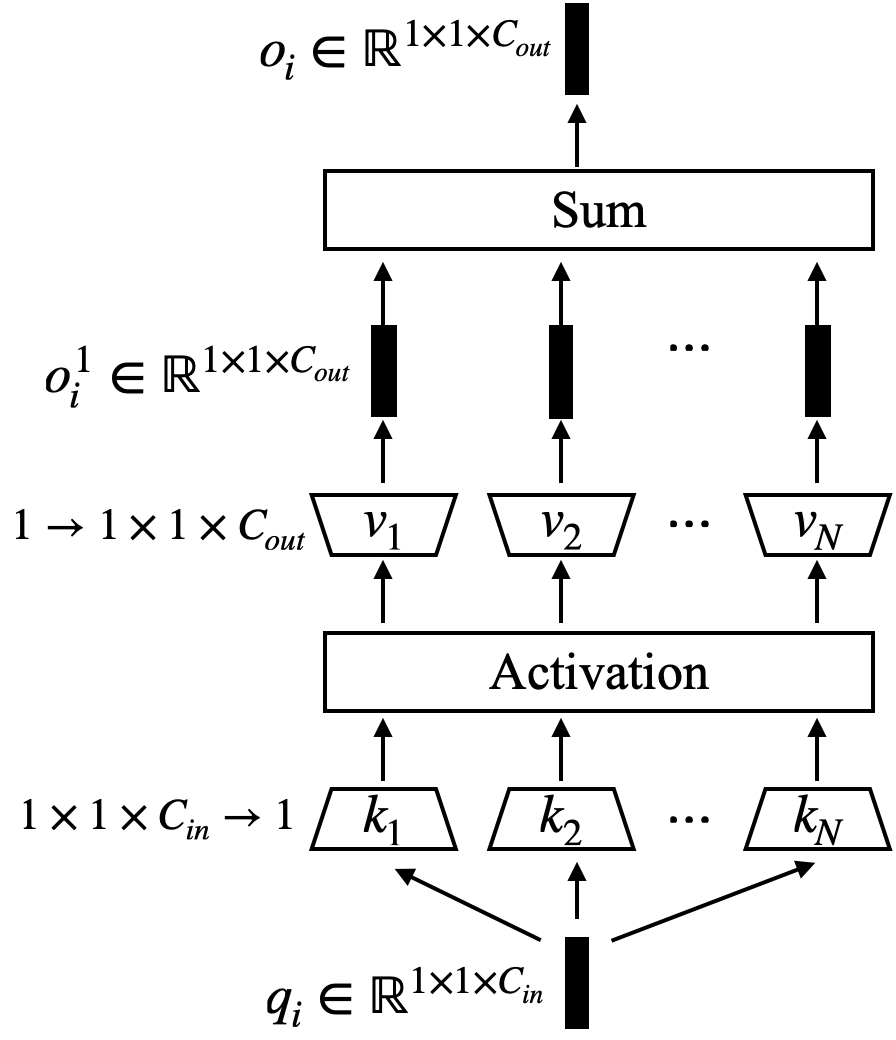}
\caption{\inlinesection{Kernel $k$ and $v$ operate in pairs in self-attention.} Each $k$ first projects $q_i$ from $C_\text{in}$-d to a scalar. After the activation function, each scalar is projected by the corresponding $v$ back to $C_\text{out}$-d. Finally, results from all $kv$ pairs are element-wise summed. As a result, self-attention achieves the $(C_{\text{in}} \rightarrow C_{\text{out}})$ projection in a very efficient way.}
\label{fig:pair}
\end{figure}

%% file: figures/fig-unify.tex
\begin{figure*}[t]
\centering
\includegraphics[width=0.97\textwidth]{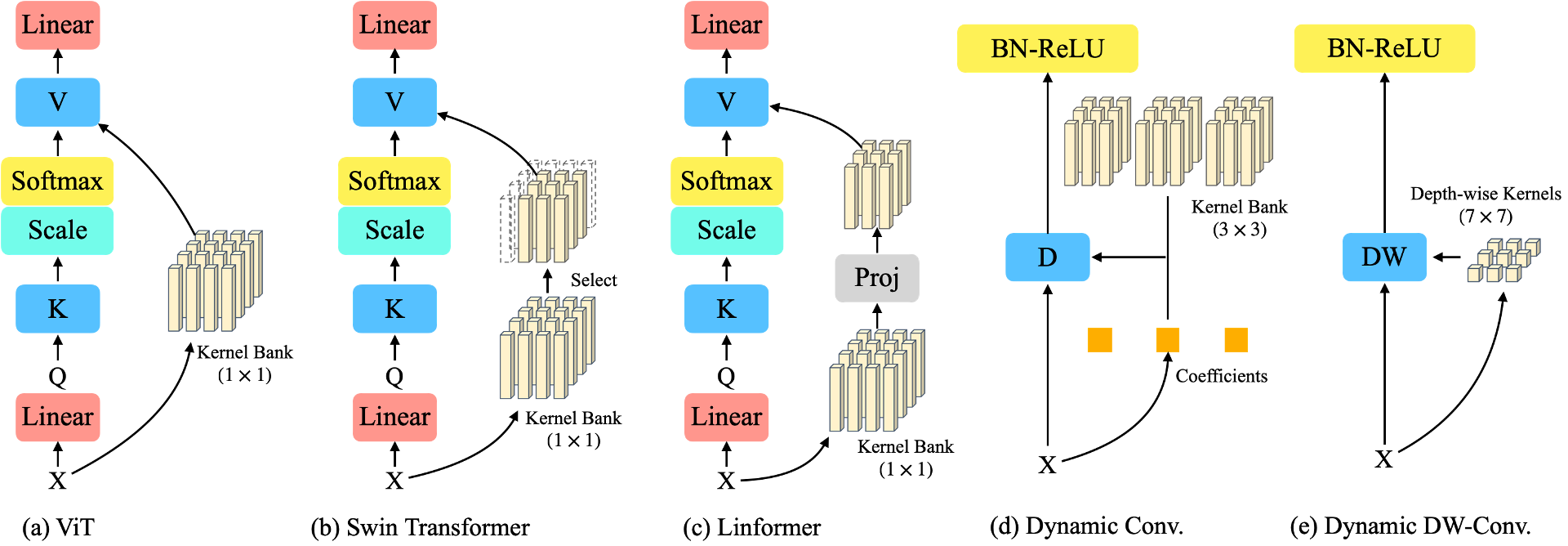}
\caption{\inlinesection{Bring existing Transformers and dynamic ConvNets into a unified framework.} Here, we show that representative global, local, and linear Transformers only differ in the way of selecting kernels from the kernel bank. In addition, since generating the weights of a standard convolutional kernel is expensive, both Transformers and ConvNets propose more efficient solutions but in quite different ways. Note that, for Transformers, we ignore the drawing of the kernel bank of keys.}
\label{fig:unify}
\end{figure*}

%% file: figures/tables/tab-unify.tex
\begin{table*}
\centering

{
\def\standard{$C \rightarrow 1 \rightarrow C$}
\caption{\inlinesection{Characterize existing Transformers and ConvNets in a unified perspective.}}
\label{tab:unify}
\begin{tabular}{l c c c c}
    \toprule
                                        & Kernel Bank   & Kernel Selection  & Kernel Type            & Kernel Size \\
    \midrule
    ViT~\cite{vit}                      & Dynamic       & Hard (all)        & \standard              & $1\times1$ \\
    Swin Transformer~\cite{swin}        & Dynamic       & Hard (local)      & \standard              & $1\times1$ \\
    Linformer~\cite{linformer}          & Dynamic       & Soft              & \standard              & $1\times1$ \\
    Dynamic Conv.~\cite{d-conv}         & Static        & Soft              & $C \rightarrow C$      & $3\times3$ \\
    Dynamic DW-Conv.~\cite{d-dw-conv}   & N/A           & N/A               & Depth-wise             & $7\times7$ \\
    \bottomrule
\end{tabular}
}
\end{table*}

%% file: sections/rethink.tex
\section{Rethink Designs of Vision Transformers}
Now that we have established a unified perspective of vision Transformers and ConvNets, we can rethink the design choices of vision Transformers in the design space of ConvNets and vice versa.
We believe such exploration can help researchers switch to a different angle during new network architecture design.
To demonstrate such potential, we provide two practical examples: (1) study the role of softmax as the activation function (Section \ref{sec:softmax}) and (2) introduce the depth-wise design into vision Transformers (Section \ref{sec:depthwise}).

\subsection{Experimental Settings}
We conduct experiments with the \emph{MMClassification} codebase~\cite{mmcls} on both ViT and Swin Transformer. We choose the ImageNet-1K~\cite{imagenet} dataset, which contains 1.28M images for training and 50K images for validation across 1K categories. And we use top-1 accuracy on the validation set as the evaluation metric. In addition, to benchmark the network efficiency, we take FLOPs, number of parameters, and activation counts (which are demonstrated by Radosavovic~\etal~\cite{regnet} to be more relevant to the throughput than FLOPs) into account. To save the computational cost, we adopt the training pipeline of DeiT~\cite{deit} for ViT. Specifically, each image is randomly resized and cropped to 224$\times$224, then augmented by random horizontal flip, RandAugment~\cite{rand-aug}, and Random Erasing~\cite{rand-erase}. Regularization methods including Mixup~\cite{mixup}, Cutmix~\cite{cutmix}, stochastic depth~\cite{stochastic}, repeated augmentation~\cite{repeat}, and Exponential Moving Average (EMA)~\cite{ema} are adopted. For optimization, we use the AdamW~\cite{adam} optimizer with the learning rate initially set to 0.001 then gradually decaying with the cosine annealing~\cite{cosine}. We train all models for 300 epochs with the batch size equal to 1024. For Swin Transformer, we follow its original training setting~\cite{swin}.

\input{figures/tables/tab-softmax}
\subsection{The Role of Softmax}
\label{sec:softmax}
Self-attention in vision Transformers adopts scaled-softmax in between $1\times1$ dynamic convolutions as the activation function, while ConvNets rarely do so. In the meanwhile, during the early stage of exploring the proper deep ConvNet architecture, activation functions, as one of the key components, have been thoroughly studied. Thus, we further ask whether these conclusions can be directly transferred to vision Transformers.

Recall that,
\begin{align}
    \text{softmax}(x)_i = \frac{\exp{(x_i / \tau)}}{\sum_{j}\exp{(x_j / \tau)}},
\end{align}
where $\tau$ is the temperature constant. Softmax serves three roles at the same time, namely normalization, channel-wise communication, and non-linearity. First, softmax normalizes the input so that the results along the channel dimension vary between 0 to 1 and sum up to 1.
Second, for the calculation at each channel position, elements at all other positions are also involved, which implicitly enables channel-wise communication. In particular, in self-attention, the channel dimension of the input to softmax denotes the key kernel index as shown in Figure~\ref{fig:pair}. Therefore, the channel-wise communication essentially becomes kernel-wise communication thus also implicitly affecting the kernel selection. Finally, as an S-shaped function, softmax provides non-linearity between two linear convolutions.

In the following, we empirically examine the significance of each role of softmax by replacing softmax with modules that are dedicated to one role and are commonly used in ConvNets. Specifically, for normalization, we substitute softmax with constant scaling or layer normalization\footnote{Layer normalization implicitly introduces non-linearity during the calculation of the standard deviation but it is not as strong as softmax or ReLU.}. And as layer normalization operates along the channel dimension, it can also serve for channel-wise communication. For non-linearity, we choose the widely-adopted ReLU function.

We choose the ViT (trained with the DeiT pipeline) and Swin Transformer as experiment objectives since they are representative works with the vanilla and local vision Transformer architectures, respectively. Note that, despite the fundamental differences between ViT and Swin Transformer being global/local attention and non-hierarchical/hierarchical, they also differ in some detailed design choices, such as the form of positional embedding (absolute \vs relative) and whether to use a class embedding for classification.

We start our exploration by removing softmax. As shown in Table~\ref{tab:softmax}, we observe that both the performances of DeiT-S and Swin-T drop (80.53 $\rightarrow$ 77.17 and 81.13 $\rightarrow$ 80.02). We then perform a simple normalization by scaling with a constant (dividing by the number of channels of each head). As a result, DeiT-S and Swin-T both obtain certain gains (77.17 $\rightarrow$ 78.15 and 80.45 $\rightarrow$ 81.15). In particular, with the constant scaling, Swin-T achieves as good a result as softmax despite the variant having no non-linearity at all. Therefore, we conclude that \textbf{normalization is important but is not limited to softmax}.

\input{figures/fig-converge}

Next, on top of the constant scaling, we further inject non-linearity with the ReLU activation. In Table~\ref{tab:softmax}, with ReLU, DeiT-S is improved from 78.15 to 80.03 while Swin-T is improved from 81.15 to 81.24.
First, the results show that ReLU is more necessary for DeiT than Swin and we conjecture the reason is that, in addition to introducing the non-linearity, ReLU also implicitly affects the kernel selection by zeroing out those kernels with negative activation, which is more important for DeiT since it has a wider range of kernels for selection.
Second, despite the effect of scaling and ReLU shows a consistent trend among ViT and Swin Transformer, compared to their corresponding softmax baselines, DeiT-S with scaling and ReLU performs worse (80.53 $\rightarrow$ 80.03) while the Swin-T variant performs slightly better (80.13 $\rightarrow$ 80.24). Therefore, before continuing to the next ablation, we study whether such differences come from detailed design choices. For this purpose, we swap the positional embedding methods between the ViT and Swin Transformer. Consequently, using the relative positional embedding, DeiT-S with scaling and ReLU obtains a very similar result with softmax (80.93 \vs 80.96). In addition, after altering to the absolute positional embedding, Swin-T also suffers from the absence of non-linearity.
Furthermore, we observe that during training, once softmax is replaced by either scaling and/or ReLU, the model usually converges faster as shown in Figure~\ref{fig:converge}. We think this observation aligns with the vanishing gradient problem of S-shaped activation functions in ConvNets, where ReLU makes the model converge faster than sigmoid. In summary, \textbf{with relative positional embedding, the softmax in ViT and Swin Transformer can be replaced with scaling-ReLU for a faster convergence rate and comparable results.}

Finally, we examine the channel-wise communication role of the softmax by replacing it with layer normalization (LN). Compared to normalizing with scaling, which performs no channel-wise communication, LN in Swin-T yields much better results. In fact, as shown in Figure~\ref{tab:softmax}, Swin-T with LN-ReLU significantly outperforms the softmax baseline (80.13 $\rightarrow$ 81.56) without any computational overhead. On DeiT-S, we first find applying LN largely degrades the performance (80.96 $\rightarrow$ 78.82), but after aligning with Swin-T where replace the class embedding of DeiT-S with global average pooling, LN-ReLU yields comparable results. We conjecture this is due to that learning a good LN is harder for ViT than Swin Transformer since the former is responsible for normalizing all kernels and the latter normalizes only selected local kernels.

\subsection{Depth-wise Vision Transformers}
\label{sec:depthwise}
\input{figures/fig-depthwise}
In Section~\ref{sec:unify}, we discuss that self-attention achieves the $(C_{\text{in}} \rightarrow C_{\text{out}})$ mapping while keeping dense channel-wise connectivity. In this section, we explore switching to sparse channel-wise connectivity, as depth-wise convolution does, to further emphasize efficiency.

When studying softmax in the previous section, we find that the non-linearity role of softmax is not always necessary. For instance, Swin-T with only scaling achieves comparable results with the softmax baseline. Moreover, since convolutions are linear operations, without any non-linear function in between, two consecutive convolutions can be merged into a single one. 
As shown in Figure~\ref{fig:depthwise}, in self-attention, with matrix multiplication, the selected kernels $k \in \mathbb{R}^{1 \times 1 \times C_{\text{in}} \times N}$ and $v \in \mathbb{R}^{1 \times 1 \times N \times C_{\text{out}}}$ can be merged into a new kernel $g \in \mathbb{R}^{1 \times 1 \times C_{\text{in}} \times C_{\text{out}}}$, which is exactly the kernel of a standard $1\times1$ convolution. This observation inspires us to replace the standard kernel with a depth-wise one.

\input{figures/tables/tab-depthwise}

For this purpose, we modify the original self-attention with the following steps. First, we discard the \emph{value} kernel bank and keep only the \emph{key} kernel bank. Next, at each spatial location, we select $N$ kernels from the bank following the same rule as self-attention (\eg, ViT selects all kernels and Swin Transformer selects local kernels) and average the selected kernels into a single one. Finally, instead of computing the dot product between the input and kernel, which projects the input to a scalar, we perform element-wise multiplication between them so that the channel dimension of the input stays unchanged. We name the modified self-attention as depth-wise self-attention since it has both the kernel bank mechanism in self-attention and the element-wise multiplication in depth-wise convolutions. And vision Transformers with the self-attention replaced to the depth-wise variant become depth-wise vision Transformers.

As shown in Table~\ref{tab:depthwise}, we conduct experiments on DeiT-Small, DeiT-Base, Swin-Tiny, and Swin-Base with the proposed depth-wise self-attention. First of all, since the depth-wise self-attention has no non-linearity, as expected, it fails on the DeiT-S. Therefore, for both ViT and Swin Transformer, by default, we employ the relative positional embedding to mitigate the non-linearity problem as discussed in Section~\ref{sec:softmax}. In particular, without softmax, in self-attention, relative positional embedding $p$ is directly applied to values:
\begin{align}
    o_{i} = (\frac{q_{i}k^\mathsf{T}}{\sqrt{C_h}}+p_{i})v = \frac{q_{i}k^\mathsf{T}}{\sqrt{C_h}}v+p_{i}v.
\end{align}
However, as the \emph{value} kernel bank is removed in the depth-wise self-attention, we apply $p$ to $k$ instead:
\begin{align}
    o_{i} = q_{i} \odot \bar{k} + p_{i}k.
\end{align}

In Table~\ref{tab:depthwise}, we show that with relative positional embedding, DeiT-S, Swin-T, and Swin-B achieve comparable results with the corresponding baselines while saving $\approx$10\% FLOPs, $\approx$8\% learnable parameters, and 20\%-40\% activation counts. 

%% file: figures/tables/tab-softmax.tex
\begin{table*}
\centering

{
\caption{\inlinesection{Ablation studies on the role of softmax as the activation function.}}
\label{tab:softmax}
\begin{tabular}{l | c c c | c c | c}
    \toprule
    Base                        & Normalization   & Non-linearity   & Channel Comm.     & Pos. Emb.     & Cls. Emb. & Top-1 Acc. \\
    \midrule
    \multirow{10}{*}{DeiT-S}    & Softmax       & Softmax           & \cmark            & Abs          &\cmark      & \textbf{80.53} \\
                                & \ph           & \ph               & \xmark            & Abs          &\cmark      & 77.17 \\
                                & Scaling       & \ph               & \xmark            & Abs          &\cmark      & 78.15 \\
                                & Scaling       & ReLU              & \xmark            & Abs          &\cmark      & 80.03 \\
                                \cmidrule{2-7}
                                & Softmax       & Softmax           & \cmark            & Rel          &\cmark      & \textbf{80.96} \\
                                & Scaling       & \ph               & \xmark            & Rel          &\cmark      & 80.45 \\
                                & Scaling       & ReLU              & \xmark            & Rel          &\cmark      & 80.93 \\
                                & LayerNorm     & \ph               & \cmark            & Rel          &\cmark      & 78.82 \\
                                & LayerNorm     & \ph               & \cmark            & Rel          &\xmark      & 80.09 \\
                                & LayerNorm     & ReLU              & \cmark            & Rel          &\xmark      & 80.64 \\
    \midrule
    \multirow{9}{*}{Swin-T}     & Softmax       & Softmax           & \cmark            & Rel          &\xmark      & 81.13 \\
                                & \ph           & \ph               & \xmark            & Rel          &\xmark      & 80.02 \\
                                & Scaling       & \ph               & \xmark            & Rel          &\xmark      & 81.15 \\
                                & Scaling       & ReLU              & \xmark            & Rel          &\xmark      & 81.24 \\
                                & LayerNorm     & \ph               & \cmark            & Rel          &\xmark      & 81.22 \\
                                & LayerNorm     & ReLU              & \cmark            & Rel          &\xmark      & \textbf{81.56} \\
                                \cmidrule{2-7}
                                & Softmax       & Softmax           & \cmark            & Abs          &\xmark      & 80.50 \\
                                & Scaling       & \ph               & \xmark            & Abs          &\xmark      & 79.99 \\
                                & Scaling       & ReLU              & \xmark            & Abs          &\xmark      & \textbf{80.57} \\
    \bottomrule
\end{tabular}
}
\end{table*}

%% file: figures/fig-converge.tex
\begin{figure}
\centering
\includegraphics[width=0.49\textwidth]{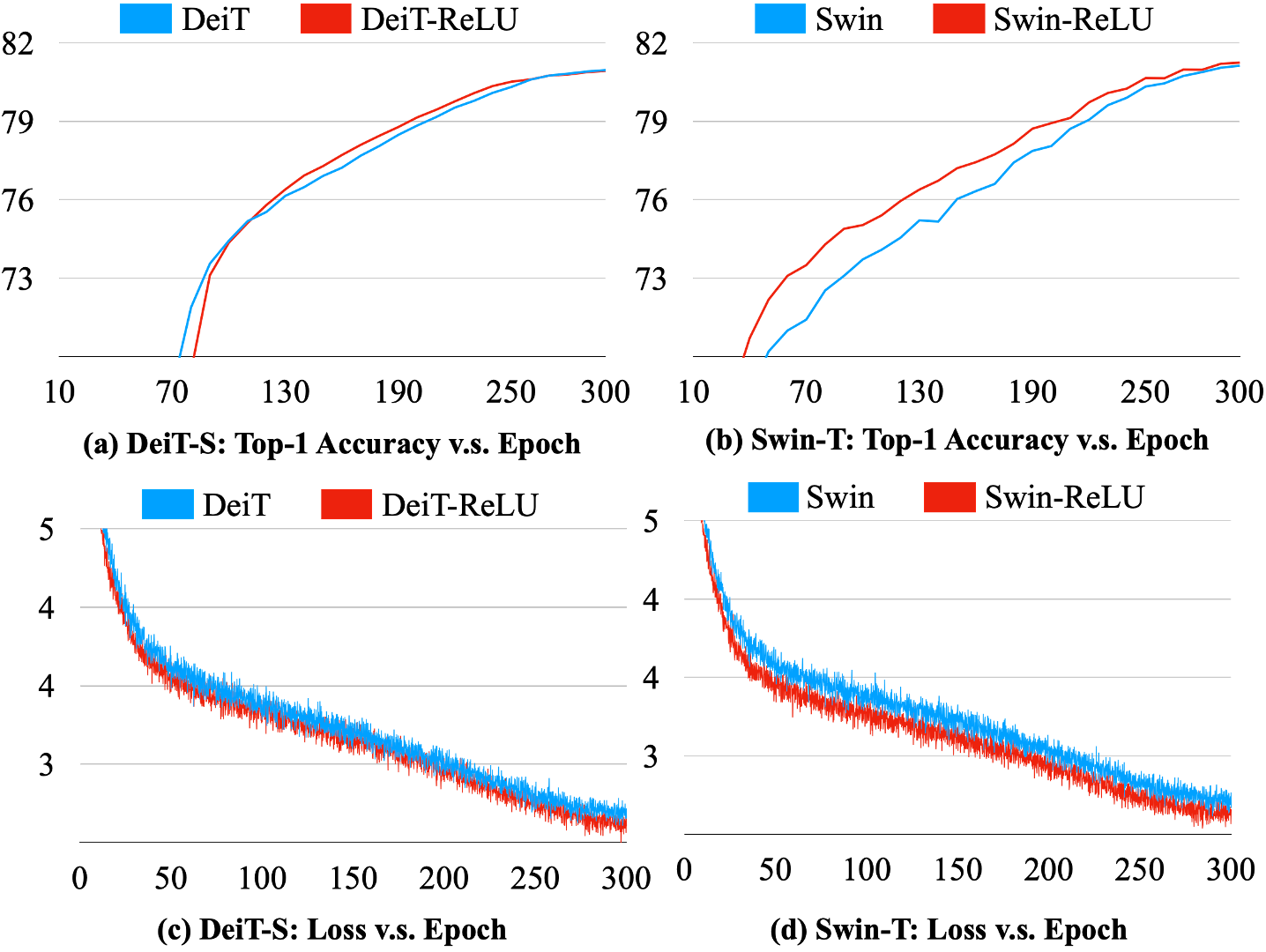}
\caption{\inlinesection{Convergence rate with different activation functions.} Here, we evaluate the convergence rate by training loss and top-1 accuracy on the validation set with respect to training epochs. Note that, DeiT in this figure adopts relative positional embeddings. Besides, all the ReLU variants use constant scaling for normalization.}
\label{fig:converge}
\vspace{-2pt}
\end{figure}

%% file: figures/fig-depthwise.tex
\begin{figure}
\centering
\includegraphics[width=0.47\textwidth]{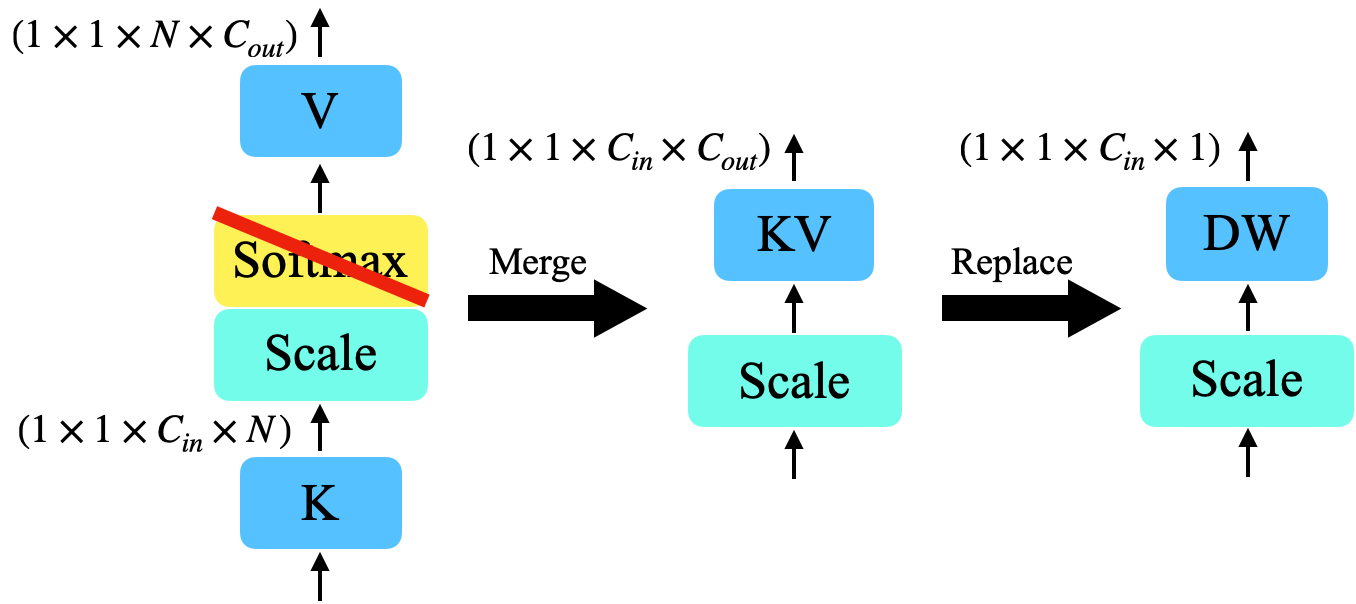}
\caption{\inlinesection{From self-attention without non-linearity to the depth-wise variant.} DW denotes depth-wise convolution. We annotate the kernel shape on the top left of each convolutional layer.}
\label{fig:depthwise}
\end{figure}

%% file: figures/tables/tab-depthwise.tex
\begin{table*}
\centering
{
\def\standard{$C \rightarrow 1 \rightarrow C$}
\caption{\inlinesection{Compare the performance of depth-wise vision Transformers with standard ones.}}
\label{tab:depthwise}
\begin{tabular}{l | c c c | c c c | c}
    \toprule
    Base
    & Act. Func.    & Kernel Type   & Pos. Emb. & GFLOPs            & MParams.          & MActs.            & Top-1 Acc. \\
    \midrule
    \multirow{6}{*}{DeiT-S}     
    & Softmax       & \standard     & Abs       & 4.608             & 22.051            & 11.947            & \textbf{80.53} \\
    & Scaling       & \standard     & Abs       & 4.366             & 22.051            & 9.448             & 78.15 \\
    & Scaling       & Depth-wise    & Abs       & \textbf{3.902}    & \textbf{20.277}   & \textbf{7.337}    & 73.61 \\
    \cmidrule{2-8}
    & Softmax       & \standard     & Rel       & 4.608             & 22.028            & 11.947            & \textbf{80.96} \\
    & Scaling       & \standard     & Rel       & 4.543             & 22.028            & 10.351            & 80.45 \\
    & Scaling       & Depth-wise    & Rel       & \textbf{4.079}    & \textbf{20.253}   & \textbf{8.241}    & 79.74 \\
    \midrule
    \multirow{3}{*}{DeiT-B}
    & Softmax       & \standard     & Abs       & 17.582            & 86.568            & 23.895            & 81.76 \\
    & Softmax       & \standard     & Rel       & 17.582            & 86.521            & 23.895            & \textbf{82.39} \\
    & Scaling       & Depth-wise    & Rel       & \textbf{15.830}   & \textbf{79.434}   & \textbf{16.481}   & 81.92 \\
    \midrule
    \multirow{3}{*}{Swin-T}     
    & Softmax       & \standard     & Rel       & 4.508             & 28.288            & 17.054            & 81.13 \\
    & Scaling       & \standard     & Rel       & 4.508             & 28.288            & 17.054            & \textbf{81.15} \\
    & Scaling       & Depth-wise    & Rel       & \textbf{4.141}    & \textbf{26.850}   & \textbf{14.112}   & 80.75 \\
    \midrule
    \multirow{3}{*}{Swin-B} 
    & Softmax       & \standard     & Rel       & 15.467            & 87.768            & 36.625            & \textbf{83.36} \\
    & Scaling       & \standard     & Rel       & 15.468            & 87.768            & 38.930            & 82.80 \\
    & Scaling       & Depth-wise    & Rel       & \textbf{14.192}   & \textbf{80.777}   & \textbf{30.256}   & 82.48 \\
    \bottomrule
\end{tabular}
}
\end{table*}

%% file: sections/discussion.tex
\section{Discussion}
\label{sec:discuss}

In this paper, we propose a new perspective that interprets vision Transformers as ConvNets with dynamic convolution through equivalently converting the self-attention block in Transformers into a combination of static convolutions, dynamic convolutions, and a non-linear activation function. In addition, we select several existing representative works from both the vision Transformer family and the dynamic ConvNet family and fit them into the proposed framework, so that we can cross-compare their design choices in a unified view. In particular, we characterize them from four dimensions, namely kernel bank, kernel selection, kernel type, and kernel size, and conduct an analysis of each dimension.
Moreover, apart from summarizing existing works, more importantly, the proposed unified perspective enables researchers to rethink design choices of vision Transformers in the design space of ConvNets and vice versa, which helps the network architecture design.
To demonstrate such potential, we provide two practical examples, including examining the role of softmax as the activation function and introducing the depth-wise design into self-attention.

\input{figures/fig-converge-supp}

\inlinesection{Future works.} We believe there is still a lot of room to explore following this direction, such as injecting spatial priors through generating $3\times3$ dynamic kernels instead $1\times1$ ones, developing self-attention-style dynamic convolutions with strides to form a hierarchy, getting rid of the heavy FFN in self-attention, and so on. We hope the proposed perspective can provide new inspiration and insights for network designs.

%% file: figures/fig-converge-supp.tex
\begin{figure*}[t]
\begin{center}
\includegraphics[width=0.98\textwidth]{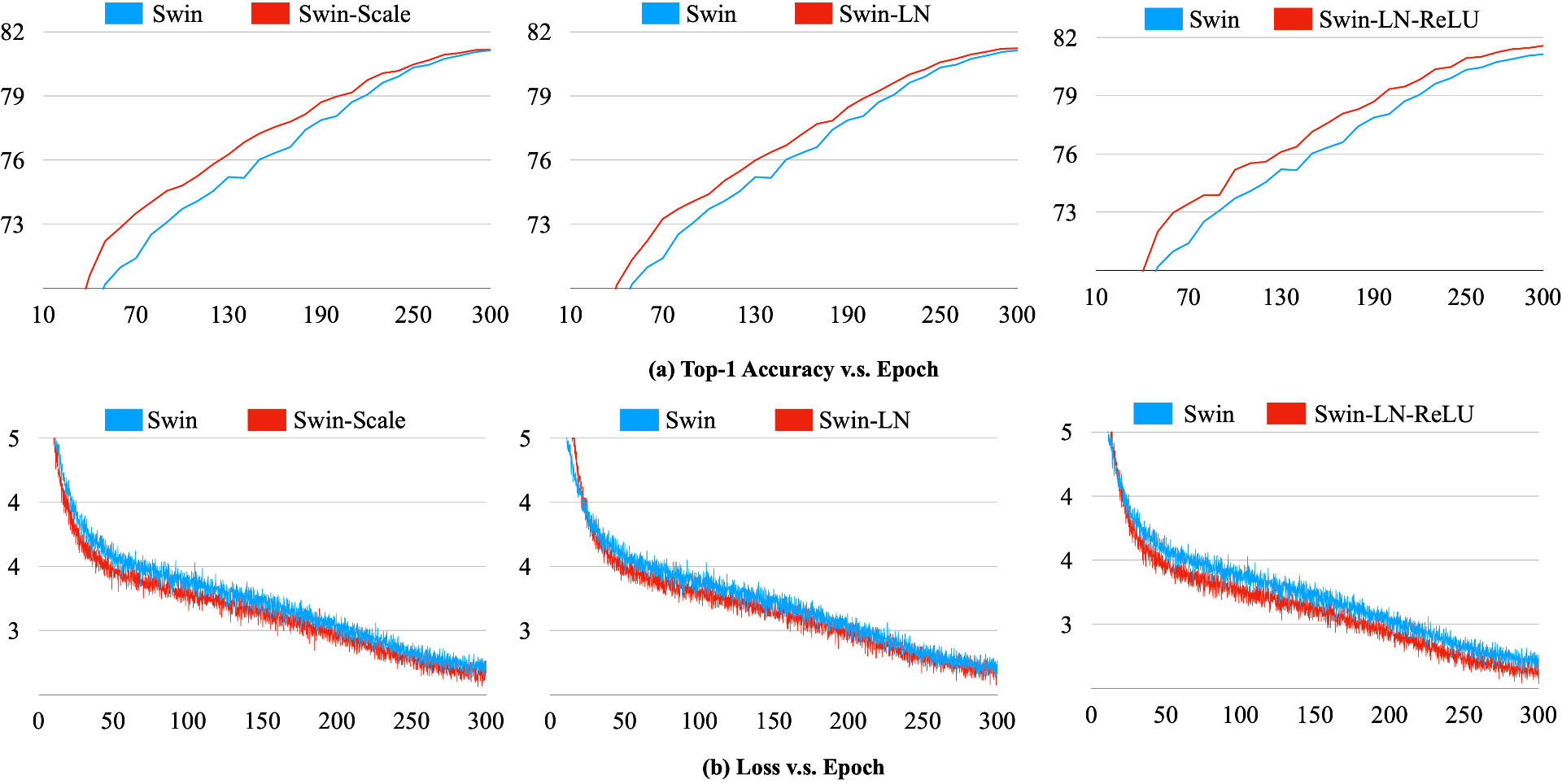}
\captionof{figure}{\inlinesection{Convergence rate with different activation functions.}}
\label{fig:converge-supp}
\end{center}
\end{figure*}

%% file: sections/supp.tex
\section{More Convergence Curves}
In our main submission, we show that when the softmax in self-attention is replaced by scaling-ReLU, despite the results being similar, the convergence rate becomes faster. In Figure~\ref{fig:converge-supp}, we provide more examples to demonstrate this point. In specific, we use constant scaling, layer normalization (LN), and LN-ReLU as the activation function, respectively. As a result, we observe the same pattern in terms of loss and accuracy as Figure~\ref{fig:converge}.

\input{figures/tables/tab-hyper-param}
\section{Implementation Details}
In Table~\ref{tab:hyper-param}, we list all the hyper-parameters and training strategies of our experiments. In general, we follow the experiment settings in the original DeiT~\cite{deit} and Swin Transformer~\cite{swin} with only a minor difference in using gradient clipping. Note that, DeiT models in different scales share the same hyper-parameters while Swin Transformers vary in the stochastic depth drop rates.

%% file: figures/tables/tab-hyper-param.tex
\begin{table}
\centering

{
\caption{\inlinesection{Training settings.}}
\label{tab:hyper-param}
\begin{tabular}{c c c}
\toprule
    Methods             & DeiT-S~\cite{deit}    & Swin-T (B)~\cite{swin} \\
\midrule
    Epochs              & 300                   & 300 \\
\midrule
    Batch size          & 1024                  & 1024 \\
    Optimizer           & AdamW                 & AdamW \\
    Learning rate       & 0.001                 & 0.001 \\
    Learning rate decay & cosine                & cosine \\
    Weight decay        & 0.05                  & 0.05 \\
    Warmup epochs       & 20                    & 20 \\
\midrule
    Label smoothing     & 0.1                   & 0.1 \\
    Dropout             & \xmark                & \xmark \\
    Stoch. Depth        & 0.1                   & 0.2 (0.5) \\
    Repeated Aug.       & \cmark                & \xmark \\
    Gradient Clip       & 5.0                   & 5.0 \\
    EMA                 & \cmark                & \xmark \\
\midrule
    Rand Augment        & 9/0.5                 & 9/0.5 \\
    Mixup prob.         & 0.8                   & 0.8 \\
    Cutmix prob.        & 1.0                   & 1.0 \\
    Erasing prob.       & 0.25                  & 0.25 \\
\bottomrule
\end{tabular}

}
\end{table}

%% file: main.bbl
\begin{thebibliography}{10}\itemsep=-1pt

\bibitem{tensor-form}
Jean-Marc Andreoli.
\newblock Convolution, attention and structure embedding.
\newblock {\em NeurIPS workshop on Graph Representation Learning}, 2019.

\bibitem{layer-norm}
Jimmy~Lei Ba, Jamie~Ryan Kiros, and Geoffrey~E Hinton.
\newblock Layer normalization.
\newblock {\em arXiv preprint arXiv:1607.06450}, 2016.

\bibitem{detr}
Nicolas Carion, Francisco Massa, Gabriel Synnaeve, Nicolas Usunier, Alexander Kirillov, and Sergey Zagoruyko.
\newblock End-to-end object detection with transformers.
\newblock In {\em ECCV}, 2020.

\bibitem{d-conv}
Yinpeng Chen, Xiyang Dai, Mengchen Liu, Dongdong Chen, Lu Yuan, and Zicheng Liu.
\newblock Dynamic convolution: Attention over convolution kernels.
\newblock In {\em CVPR}, 2020.

\bibitem{mmcls}
MMClassification Contributors.
\newblock Openmmlab's image classification toolbox and benchmark.
\newblock \url{https://github.com/open-mmlab/mmclassification}, 2020.

\bibitem{relation}
Jean-Baptiste Cordonnier, Andreas Loukas, and Martin Jaggi.
\newblock On the relationship between self-attention and convolutional layers.
\newblock In {\em ICLR}, 2020.

\bibitem{rand-aug}
Ekin~D Cubuk, Barret Zoph, Jonathon Shlens, and Quoc~V Le.
\newblock Randaugment: Practical automated data augmentation with a reduced search space.
\newblock In {\em CVPRW}, 2020.

\bibitem{imagenet}
Jia Deng, Wei Dong, Richard Socher, Li-Jia Li, Kai Li, and Li Fei-Fei.
\newblock Imagenet: A large-scale hierarchical image database.
\newblock In {\em CVPR}, 2009.

\bibitem{RepLKNet}
Xiaohan Ding, Xiangyu Zhang, Jungong Han, and Guiguang Ding.
\newblock Scaling up your kernels to 31x31: Revisiting large kernel design in cnns.
\newblock In {\em CVPR}, 2022.

\bibitem{vit}
Alexey Dosovitskiy, Lucas Beyer, Alexander Kolesnikov, Dirk Weissenborn, Xiaohua Zhai, Thomas Unterthiner, Mostafa Dehghani, Matthias Minderer, Georg Heigold, Sylvain Gelly, Jakob Uszkoreit, and Neil Houlsby.
\newblock An image is worth 16x16 words: Transformers for image recognition at scale.
\newblock In {\em ICLR}, 2021.

\bibitem{d-dw-conv}
Qi Han, Zejia Fan, Qi Dai, Lei Sun, Ming-Ming Cheng, Jiaying Liu, and Jingdong Wang.
\newblock On the connection between local attention and dynamic depth-wise convolution.
\newblock In {\em ICLR}, 2021.

\bibitem{resnet}
Kaiming He, Xiangyu Zhang, Shaoqing Ren, and Jian Sun.
\newblock Deep residual learning for image recognition.
\newblock In {\em CVPR}, 2016.

\bibitem{repeat}
Elad Hoffer, Tal Ben-Nun, Itay Hubara, Niv Giladi, Torsten Hoefler, and Daniel Soudry.
\newblock Augment your batch: Improving generalization through instance repetition.
\newblock In {\em CVPR}, 2020.

\bibitem{genet}
Jie Hu, Li Shen, Samuel Albanie, Gang Sun, and Andrea Vedaldi.
\newblock Gather-excite: Exploiting feature context in convolutional neural networks.
\newblock {\em NeurIPS}, 2018.

\bibitem{senet}
Jie Hu, Li Shen, and Gang Sun.
\newblock Squeeze-and-excitation networks.
\newblock In {\em CVPR}, 2018.

\bibitem{stochastic}
Gao Huang, Yu Sun, Zhuang Liu, Daniel Sedra, and Kilian~Q Weinberger.
\newblock Deep networks with stochastic depth.
\newblock In {\em ECCV}, 2016.

\bibitem{filter}
Xu Jia, Bert De~Brabandere, Tinne Tuytelaars, and Luc~V Gool.
\newblock Dynamic filter networks.
\newblock {\em NeurIPS}, 2016.

\bibitem{adam}
Diederik~P Kingma and Jimmy Ba.
\newblock Adam: A method for stochastic optimization.
\newblock {\em arXiv preprint arXiv:1412.6980}, 2014.

\bibitem{alexnet}
Alex Krizhevsky, Ilya Sutskever, and Geoffrey~E Hinton.
\newblock Imagenet classification with deep convolutional neural networks.
\newblock {\em Communications of the ACM}, 2017.

\bibitem{involution}
Duo Li, Jie Hu, Changhu Wang, Xiangtai Li, Qi She, Lei Zhu, Tong Zhang, and Qifeng Chen.
\newblock Involution: Inverting the inherence of convolution for visual recognition.
\newblock In {\em CVPR}, 2021.

\bibitem{swin}
Ze Liu, Yutong Lin, Yue Cao, Han Hu, Yixuan Wei, Zheng Zhang, Stephen Lin, and Baining Guo.
\newblock Swin transformer: Hierarchical vision transformer using shifted windows.
\newblock In {\em ICCV}, 2021.

\bibitem{convnext}
Zhuang Liu, Hanzi Mao, Chao-Yuan Wu, Christoph Feichtenhofer, Trevor Darrell, and Saining Xie.
\newblock A convnet for the 2020s.
\newblock In {\em CVPR}, 2022.

\bibitem{cosine}
Ilya Loshchilov and Frank Hutter.
\newblock Sgdr: Stochastic gradient descent with warm restarts.
\newblock In {\em ICLR}, 2017.

\bibitem{relu}
Vinod Nair and Geoffrey~E Hinton.
\newblock Rectified linear units improve restricted boltzmann machines.
\newblock In {\em ICML}, 2010.

\bibitem{integration}
Xuran Pan, Chunjiang Ge, Rui Lu, Shiji Song, Guanfu Chen, Zeyi Huang, and Gao Huang.
\newblock On the integration of self-attention and convolution.
\newblock In {\em CVPR}, 2022.

\bibitem{ema}
Boris~T Polyak and Anatoli~B Juditsky.
\newblock Acceleration of stochastic approximation by averaging.
\newblock {\em SIAM journal on control and optimization}, 1992.

\bibitem{regnet}
Ilija Radosavovic, Raj~Prateek Kosaraju, Ross Girshick, Kaiming He, and Piotr Doll{\'a}r.
\newblock Designing network design spaces.
\newblock In {\em CVPR}, 2020.

\bibitem{vggnet}
Karen Simonyan and Andrew Zisserman.
\newblock Very deep convolutional networks for large-scale image recognition.
\newblock In {\em ICLR}, 2014.

\bibitem{nlp2}
Yi Tay, Mostafa Dehghani, Jai~Prakash Gupta, Vamsi Aribandi, Dara Bahri, Zhen Qin, and Donald Metzler.
\newblock Are pretrained convolutions better than pretrained transformers?
\newblock In {\em ACL/IJCNLP}, 2021.

\bibitem{deit}
Hugo Touvron, Matthieu Cord, Matthijs Douze, Francisco Massa, Alexandre Sablayrolles, and Herv{\'e} J{\'e}gou.
\newblock Training data-efficient image transformers \& distillation through attention.
\newblock In {\em ICML}, 2021.

\bibitem{transformer}
Ashish Vaswani, Noam Shazeer, Niki Parmar, Jakob Uszkoreit, Llion Jones, Aidan~N Gomez, {\L}ukasz Kaiser, and Illia Polosukhin.
\newblock Attention is all you need.
\newblock {\em NeurIPS}, 2017.

\bibitem{linformer}
Sinong Wang, Belinda~Z Li, Madian Khabsa, Han Fang, and Hao Ma.
\newblock Linformer: Self-attention with linear complexity.
\newblock {\em arXiv preprint arXiv:2006.04768}, 2020.

\bibitem{pyramid}
Wenhai Wang, Enze Xie, Xiang Li, Deng-Ping Fan, Kaitao Song, Ding Liang, Tong Lu, Ping Luo, and Ling Shao.
\newblock Pyramid vision transformer: A versatile backbone for dense prediction without convolutions.
\newblock In {\em ICCV}, 2021.

\bibitem{nlp1}
Felix Wu, Angela Fan, Alexei Baevski, Yann Dauphin, and Michael Auli.
\newblock Pay less attention with lightweight and dynamic convolutions.
\newblock In {\em ICLR}, 2019.

\bibitem{cvt}
Haiping Wu, Bin Xiao, Noel Codella, Mengchen Liu, Xiyang Dai, Lu Yuan, and Lei Zhang.
\newblock Cvt: Introducing convolutions to vision transformers.
\newblock In {\em ICCV}, 2021.

\bibitem{ceit}
Kun Yuan, Shaopeng Guo, Ziwei Liu, Aojun Zhou, Fengwei Yu, and Wei Wu.
\newblock Incorporating convolution designs into visual transformers.
\newblock In {\em ICCV}, 2021.

\bibitem{cutmix}
Sangdoo Yun, Dongyoon Han, Seong~Joon Oh, Sanghyuk Chun, Junsuk Choe, and Youngjoon Yoo.
\newblock Cutmix: Regularization strategy to train strong classifiers with localizable features.
\newblock In {\em ICCV}, 2019.

\bibitem{mixup}
Hongyi Zhang, Moustapha Cisse, Yann~N Dauphin, and David Lopez-Paz.
\newblock Mixup: Beyond empirical risk minimization.
\newblock In {\em ICLR}, 2018.

\bibitem{setr}
Sixiao Zheng, Jiachen Lu, Hengshuang Zhao, Xiatian Zhu, Zekun Luo, Yabiao Wang, Yanwei Fu, Jianfeng Feng, Tao Xiang, Philip~H.S. Torr, and Li Zhang.
\newblock Rethinking semantic segmentation from a sequence-to-sequence perspective with transformers.
\newblock In {\em CVPR}, 2021.

\bibitem{rand-erase}
Zhun Zhong, Liang Zheng, Guoliang Kang, Shaozi Li, and Yi Yang.
\newblock Random erasing data augmentation.
\newblock In {\em AAAI}, 2020.

\end{thebibliography}
